\documentclass[11pt,a4paper]{article}
\usepackage[hyperref]{acl/acl2019}
\usepackage{times}
\usepackage{latexsym}
\usepackage{graphicx}
\usepackage{amssymb}
\usepackage{multirow}
\usepackage[flushleft]{threeparttable}
\usepackage{url}
\usepackage{floatrow}
\usepackage{wrapfig}
\usepackage[capitalise]{cleveref}
\usepackage{booktabs}
\usepackage{comment}
\newfloatcommand{capbtabbox}{table}[][\FBwidth]
\newcommand{\mysubsection}[1]{\vspace{0.3em}\noindent\textbf{#1}}

\usepackage{pifont}
\newcommand{\xmark}{x}

\aclfinalcopy %

\makeatletter
\def\hlinewd#1{%
  \noalign{\ifnum0=`}\fi\hrule \@height #1 \futurelet
   \reserved@a\@xhline}
\makeatother

\title{Identifying Visible Actions in Lifestyle Vlogs}

\author{Oana Ignat\textsuperscript{1},
Laura Burdick\textsuperscript{1},
  Jia Deng\textsuperscript{2},
  Rada Mihalcea\textsuperscript{1} \\
  \textsuperscript{1}University of Michigan,
  \textsuperscript{2}Princeton University\\
  \texttt{\{oignat,wenlaura,mihalcea\}@umich.edu},  \texttt{jiadeng@cs.princeton.edu}
  }

\date{}

\begin{document}
\maketitle

\begin{abstract}
We consider the task of identifying human actions visible in online videos. We focus on the widely spread genre of lifestyle vlogs, which consist of videos of people performing actions while verbally describing them. Our goal is to identify if actions mentioned in the speech description of a video are visually present. We construct a dataset with crowdsourced manual annotations of visible actions, and introduce a multimodal algorithm that  leverages information derived from visual and linguistic clues to automatically infer which actions are visible in a video. We demonstrate that our multimodal algorithm outperforms algorithms based only on one  modality at a time.
\end{abstract}

\begin{table*}\setlength{\tabcolsep}{3pt}
    \centering
    \begin{tabular}{l c c c c c c c}
    \toprule
        Dataset  & \#Actions & \#Verbs& \#Actors & Implicit & Label types\\
        \midrule
       Ours & 4340 & 580 & 10  & \checkmark & \checkmark  \\
        \midrule
        VLOG  \small{\cite{fouhey2018lifestyle}} &- & - & 10.7k  & \checkmark & \checkmark \\
        Kinetics \small{\cite{kay2017kinetics}}&600& 270 & - & \xmark & \xmark \\
        ActivityNet \small{\cite{caba2015activitynet}} &203 & - & -  & \xmark & \xmark \\
        MIT \small{\cite{monfort2019moments}}&339& 339 & - & \xmark & \xmark \\
        AVA \small{\cite{gu2018ava}}&80&80 & 192  & \checkmark & \xmark \\
        Charades \small{\cite{sigurdsson2016hollywood}}&157 & 30 &267  & \xmark & \xmark \\
        MPII Cooking \small{\cite{rohrbach2012database}}&78 &78 & 12 & \checkmark & \xmark \\
        \bottomrule
    \end{tabular}
    \caption{Comparison between our dataset and other video human action recognition datasets. \# Actions show either the number of action classes in that dataset (for the other datasets), or the number of unique visible actions in that dataset (ours); \# Verbs shows the number of unique verbs in the actions; Implicit is the type of data gathering method (versus explicit); Label types are either post-defined (first gathering data and then annotating actions): \checkmark, or pre-defined (annotating actions before gathering data): \xmark.}
    \label{tab:comparison_statistics}
    
\end{table*}

\section{Introduction}

There has been a surge of recent interest in detecting human actions in videos. Work in this space has mainly focused on learning actions from articulated human pose \cite{du2015hierarchical,vemulapalli2014human,zhang2017view} or mining spatial and temporal information from videos \cite{simonyan2014two,wang2016temporal}. A number of resources have been produced, including Action Bank \cite{sadanand2012action}, NTU RGB+D \cite{shahroudy2016ntu}, SBU Kinect Interaction \cite{yun2012two}, and PKU-MMD \cite{liu2017pku}.

Most research on video action detection has gathered video information for a set of pre-defined actions \cite{caba2015activitynet,real2017youtube,kay2017kinetics}, an approach known as {\it explicit data gathering} \cite{fouhey2018lifestyle}. For instance, given an action such as ``open door,'' a system would identify videos that include a visual depiction of this action. While this approach is able to detect a specific set of actions, whose choice may be guided by downstream applications, it  achieves high precision at the cost of low recall. In many cases, the set of predefined actions is small (e.g., 203 activity classes in \citealt{caba2015activitynet}), and for some actions, the number of visual depictions is very small. 

An alternative approach is to start with a set of videos, and identify all the actions present in these videos \cite{damen2018scaling,bregler1997learning}. This approach has been referred to as {\it implicit data gathering,} and it typically leads to the identification of a larger number of actions, possibly with a small number of examples per action.

In this paper, we use an implicit data gathering approach to label human activities in videos. To the best of our knowledge, we are the first to explore video action recognition using both transcribed audio and video information. We focus on the popular genre of lifestyle vlogs, which consist of videos of people demonstrating routine actions while verbally describing them. We use these videos to develop methods to identify if actions are visually present.

The paper makes three main contributions. First, we introduce a novel dataset consisting of 1,268 short video clips paired with sets of actions mentioned in the video transcripts, as well as manual annotations of whether the actions are visible or not. The dataset includes a total of 14,769 actions, 4,340 of which are visible. Second, we propose a set of strong baselines to determine whether an action is visible or not. Third, we introduce a multimodal neural architecture that combines information drawn from visual and linguistic clues, and show that it improves over models that rely on one modality at a time.

By making progress towards automatic action recognition, in addition to contributing to video understanding, this work has a number of important and exciting applications, including sports analytics \cite{fani2017hockey}, human-computer interaction \cite{rautaray2015vision}, and automatic analysis of surveillance video footage \cite{ji20123d}.

The paper is organized as follows. We begin by discussing related work, then describe our data collection and annotation process. We next overview our experimental set-up and introduce a multimodal method for identifying visible actions in videos. Finally, we discuss our results and conclude with general directions for future work.

\section{Related Work}

There has been substantial work on action recognition in the computer vision community, focusing on creating datasets \cite{soomro2012ucf101, karpathy2014large, sigurdsson2016hollywood, caba2015activitynet} or introducing new methods \cite{herath2017going, carreira2017quo, donahue2015long, tran2015learning}.  Table \ref{tab:comparison_statistics} compares our dataset with previous action recognition datasets.\footnote{Note that the number of actions shown for our dataset reflects the number of unique visible actions in the dataset and not the number of action classes, as in other datasets. This is due to our annotation process (see \S \ref{sec:dataCollection}).}

The largest datasets that have been compiled to date are based on YouTube videos  \cite{caba2015activitynet,real2017youtube,kay2017kinetics}. These actions cover a broad range of classes including human-object interactions such as cooking \cite{rohrbach2014coherent,das2013thousand,rohrbach2012database} and playing tennis \cite{karpathy2014large}, as well as human-human interactions such as shaking hands and hugging \cite{gu2018ava}.

Similar to our work, some of these previous datasets have considered everyday routine actions \cite{caba2015activitynet,real2017youtube,kay2017kinetics}. However, because these  datasets rely on videos uploaded on YouTube, it has been observed they can be  potentially biased towards unusual situations \cite{kay2017kinetics}. For example, searching for videos with the query ``drinking tea" results mainly in unusual videos such as dogs or birds drinking tea. This bias can be addressed by paying people to act out everyday scenarios \cite{sigurdsson2016hollywood}, but this  can end up being very expensive. In our work, we address this bias by changing the  approach used to search for videos. Instead of searching for actions in an explicit way, using queries such as ``opening a fridge'' or ``making the bed,'' we search for more general videos using queries such as ``my morning routine.'' %
This approach has been referred to as implicit (as opposed to explicit) data gathering, and was shown to result in a greater number of videos with more realistic action depictions \cite{fouhey2018lifestyle}.

Although we use implicit data gathering as proposed in the past, unlike \cite{fouhey2018lifestyle} and other human action recognition datasets, we search for routine videos that contain rich audio descriptions of the actions being performed, and we use this transcribed audio to extract actions. In these lifestyle vlogs, a vlogger typically performs an action while also describing it in detail. To the best of our knowledge, we are the first to build a video action recognition dataset using both transcribed audio and video information.

Another important difference between our methodology and previously proposed methods is that we extract action labels from the transcripts. By gathering data before annotating the actions, our action labels are post-defined (as in \citealt{fouhey2018lifestyle}). This is unlike the majority of the existing human action datasets that use pre-defined labels \cite{sigurdsson2016hollywood,caba2015activitynet,real2017youtube,kay2017kinetics, gu2018ava, das2013thousand,rohrbach2012database, monfort2019moments}.  Post-defined labels allow us to use a larger set of labels, expanding on the simplified label set used in earlier datasets. These action labels are more inline with  everyday scenarios, where people often use different names for the same action. For example, when interacting with a robot, a user could refer to an action in a variety of ways; our dataset includes the actions ``stick it into the freezer,'' ``freeze it,'' ``pop into the freezer,'' and ``put into the freezer,'' variations, which would not be included in current human action recognition datasets.

In addition to human action recognition, our work relates to other multimodal tasks such as visual question answering \cite{jang2017tgif,wu2017visual}, video summarization \cite{gygli2014creating, song2015tvsum}, and mapping text descriptions to video content \cite{karpathy2015deep, rohrbach2016grounding}. Specifically, we use an  architecture similar to \cite{jang2017tgif}, where an LSTM \cite{hochreiter1997long} is used together with frame-level  visual features such as  Inception \cite{szegedy2016rethinking}, and sequence-level features such as C3D \cite{tran2015learning}. However, unlike \cite{jang2017tgif} who encode the textual information (question-answers pairs) using an LSTM, we chose instead to encode our textual information (action descriptions and their contexts) using a large-scale language model ELMo \cite{peters2018deep}. 

Similar to previous research on multimodal methods \cite{lei2018tvqa, xu2015semantic,wu2013realistic, jang2017tgif}, we also perform feature ablation to determine the role played by each modality in solving the task. Consistent with earlier work, we observe that the textual modality leads to the highest performance across individual modalities, and that the multimodal model combining textual and visual clues has the best overall performance.

\section{Data Collection and Annotation}\label{sec:dataCollection}
We collect a dataset of routine and do-it-yourself (DIY) videos from YouTube, consisting of  people performing daily activities, such as making breakfast or cleaning the house. These videos also typically include a detailed verbal description of the actions being depicted. We choose to focus on these lifestyle vlogs because they are very popular, with tens of millions having been uploaded on YouTube; \cref{tab:nbresults_search_queries} shows the approximate number of videos available for several routine queries. Vlogs also capture a wide range of everyday activities; on average, we find thirty different visible human actions in five minutes of video.

By collecting routine videos, instead of searching explicitly for actions, we do {\it implicit} data gathering, a form of data collection introduced by \citealt{fouhey2018lifestyle}. Because everyday actions are common and not unusual, searching for them directly does not return many results. In contrast, by collecting routine videos, we find many everyday activities present in these videos. %

\begin{table}
    \centering
    \scalebox{0.9}{
    \begin{tabular}{lc}
    \hline
    Query&Results\\
    \midrule
    my morning routine & 28M+ \\
    my after school routine & 13M+ \\
    my workout routine & 23M+ \\
    my cleaning routine & 13M+ \\
    DIY & 78M+ \\ 
    \hline
    \end{tabular}
    }
    \caption{Approximate number of videos found when searching for routine and do-it-yourself queries on YouTube.}
    \label{tab:nbresults_search_queries}
\end{table}

\subsection{Data Gathering}
\begin{figure}[!htb]
    \centering
    \includegraphics[width=\textwidth]{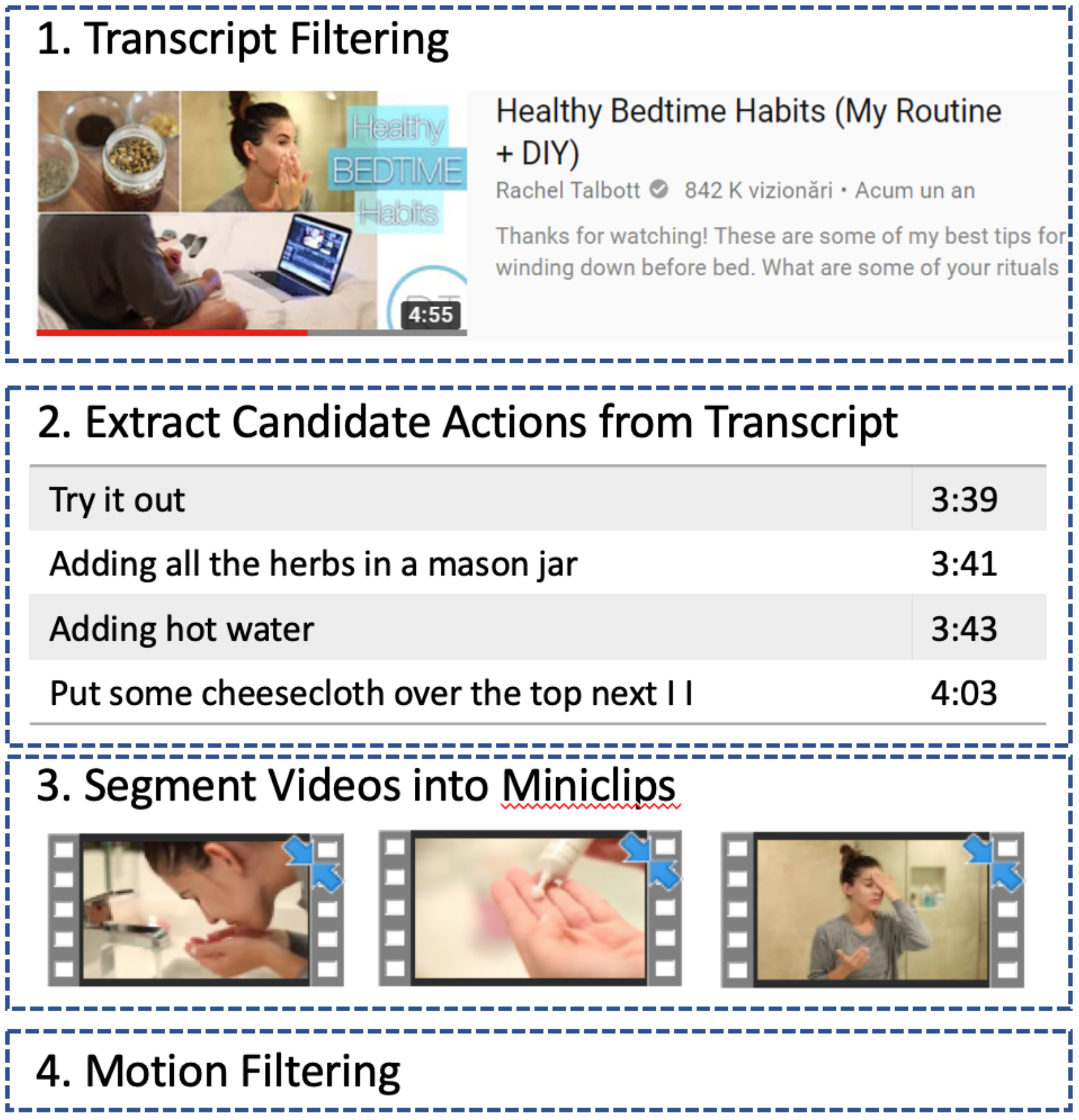}
    \caption{Overview of the data gathering pipeline.}
    \label{fig:dataGathering}
\end{figure}

We build a data gathering pipeline (see Figure \ref{fig:dataGathering}) to automatically extract and filter videos and their transcripts from YouTube. The input to the pipeline is manually selected YouTube channels. Ten channels are chosen for their rich routine videos, where the actor(s) describe their actions in great detail. From each channel, we manually select two different playlists, and from each playlist, we randomly download ten videos.

The following data processing steps are applied: 

\mysubsection{Transcript Filtering.} 
Transcripts are automatically generated by YouTube.
We filter out videos that do not contain any transcripts or that contain transcripts with an average (over the entire video) of less than 0.5 words per second.
These videos do not contain detailed action descriptions so we cannot effectively leverage textual information. 

\mysubsection{Extract Candidate Actions from Transcript.}
Starting with the transcript, we generate a noisy list of potential actions. This is done using the Stanford parser \cite{chen2014fast} to split the transcript into sentences and identify verb phrases, augmented by a set of hand-crafted rules to eliminate some parsing errors. The resulting actions are noisy, containing phrases such as ``found it helpful if you'' and ``created before up the top you.'' 

\mysubsection{Segment Videos into Miniclips.} The length of our collected videos varies from two minutes to twenty minutes. To ease the annotation process, we split each video into miniclips (short video sequences of maximum one minute). Miniclips are split to minimize the chance that the same action is shown across multiple miniclips. This is done automatically, based on the transcript timestamp of each action.
Because YouTube transcripts have timing information, we are able to line up each action with its corresponding frames in the video. We sometimes notice a gap of several seconds between the time an action occurs in the transcript and the time it is shown in the video. To address this misalignment, we first map the actions to the miniclips using the time information from the transcript. We then expand the miniclip by 15 seconds before the first action and 15 seconds after the last action. This increases the chance that all actions will be captured in the miniclip.

\mysubsection{Motion Filtering.} We remove miniclips that do not contain much movement. We sample one out of every one hundred frames of the miniclip, and compute the 2D correlation coefficient between these sampled frames. If the median of the obtained values is greater than a certain threshold (we choose 0.8), we filter out the miniclip. %
Videos with low movement tend to show people sitting in front of the camera, describing their routine, but not acting out what they are saying. There can be many actions in the transcript, but if they are not depicted in the video, we cannot leverage the video information.

\begin{figure*}
\begin{floatrow}
\ffigbox[\FBwidth]{
    \includegraphics[height=6cm]{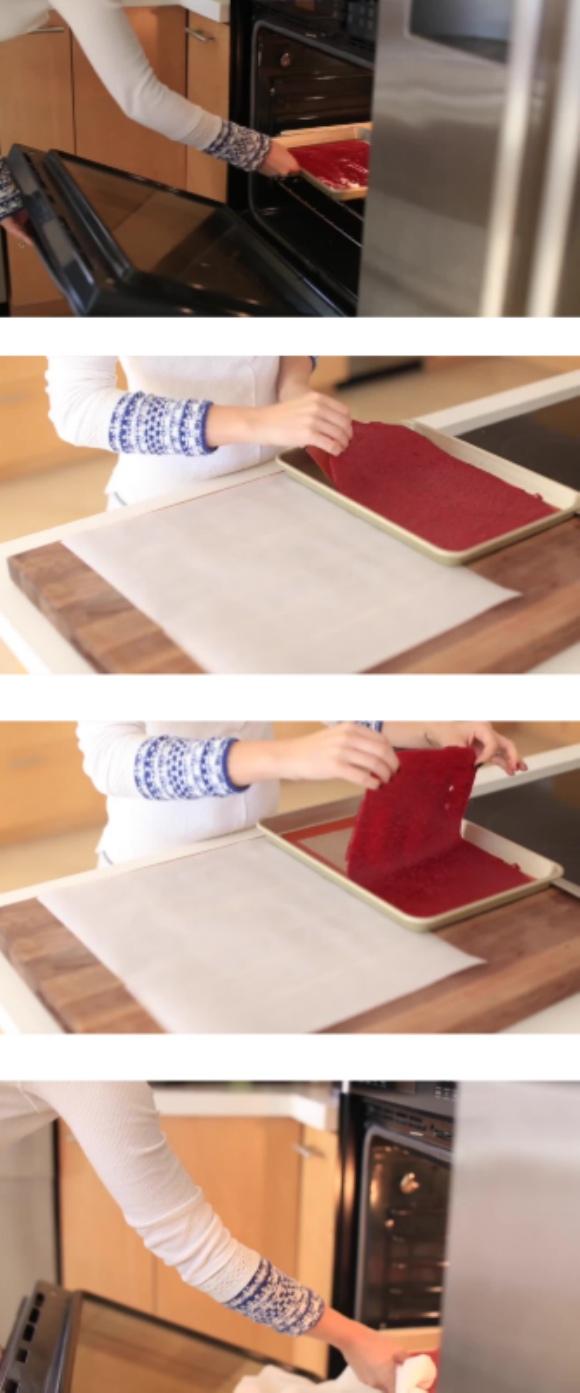}
}{
\caption*{}
}
\capbtabbox{%
\scalebox{0.9}{
  \begin{tabular}{l}
        ...\\
03:24 you're gonna actually cook it \\
03:27 and it you're gonna bake it for \\
03:30 about six hours it's definitely a \\
03:32 long time so keep in mind that it's \\
03:34 basically just dehydrating it \\
03:50 after what seems like an eternity in \\
03:53 the oven you're going to take it out \\
03:55 it's actually dehydrated at that point \\
03:57 which is fabulous because you can\\
03:59 pull it right off the baking sheet and\\ 
04:01 you're going to put it on to some\\
04:03 parchment paper and then you're \\
...\\

    \end{tabular}
    }
}{%
    \caption*{}
}
\capbtabbox{%
\scalebox{0.9}{
  \begin{tabular}{l|c}
  \hline
    Action&Visible?\\
    \midrule
    actually cook it&\checkmark\\
bake it for&\checkmark\\
take it out&\checkmark\\
pull it right off &\checkmark\\
\hspace{0.3cm}the baking sheet&\\
put it on to some &\checkmark\\
\hspace{0.3cm}parchment paper&\\
\midrule
so keep in mind that &\xmark\\
seems like an eternity &\xmark\\
\hspace{0.3cm}in the oven&\\
dehydrated at that &\xmark\\
\hspace{0.3cm}point which&\\
\hline 
    \end{tabular}
    }
}{%
    \caption*{}
}
\end{floatrow}

\begin{floatrow}
    \ffigbox[\FBwidth]{
    \hspace{15.5cm}
    
    }{
        \caption{Sample video frames, transcript, and annotations.}
        \label{fig:sample}
    }
    
\end{floatrow}
\end{figure*}
\subsection{Visual Action Annotation}

Our goal is to identify which of the actions extracted from the transcripts are visually depicted in the videos. We create an annotation task on Amazon Mechanical Turk (AMT) to identify actions that are visible. 

We give each AMT turker a HIT consisting of five miniclips with up to seven actions generated from each miniclip.
The turker is asked to assign a label (\textit{visible} in the video; \textit{not visible} in the video; \textit{not an action}) to each action.
Because it is difficult to reliably separate \textit{not visible} and \textit{not an action}, we group these labels together. 

Each miniclip is annotated by three different turkers.  For the final annotation, we use the label assigned by the majority of turkers, i.e., \textit{visible} or \textit{not visible / not an action}.

To help detect spam, we identify and reject the turkers that assign the same label for every action in all five miniclips that they annotate. Additionally, each HIT contains a ground truth miniclip that has been pre-labeled by two reliable annotators. Each ground truth miniclip has more than four actions with labels that were agreed upon by both reliable annotators. 
We compute accuracy between a turker's answers and the ground truth annotations; if this accuracy is less than 20\%, we reject the HIT as spam. 

\begin{figure}[ht]
\begin{floatrow}
\capbtabbox{%
\scalebox{0.95}{
  \begin{tabular}{l|c c c c}
 \hline
        Action &  \#1 &  \#2 &  \#3 &  GT\\
        \midrule
        make sure your skin& \xmark & \xmark & \checkmark & \xmark \\
        cleansed before you& \checkmark & \xmark & \checkmark & \checkmark\\
        do all that& \xmark & \xmark & \checkmark & \xmark \\
        absorbing all that & \xmark &  \xmark & \checkmark & \xmark \\
        \hspace{0.3cm}serum when there&&&&\\
        move on& \xmark & \xmark & \xmark & \xmark \\
\hline 
    \end{tabular}
    }
}{%
    \caption*{}
}
\end{floatrow}

\begin{floatrow}
\ffigbox[\FBwidth]{%

  \includegraphics[width=\textwidth]{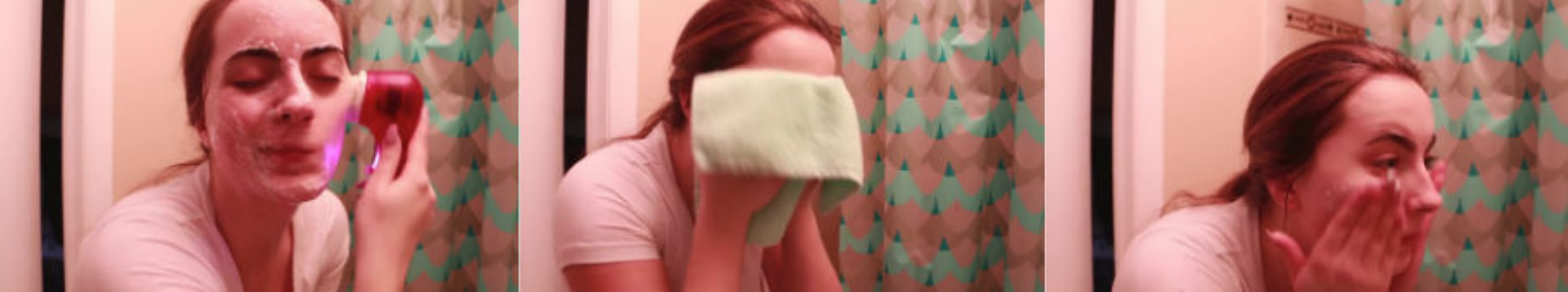}
}{%
    \caption*{}
}
\end{floatrow}
    \caption{An example of low agreement. The table shows actions and annotations from workers \#1, \#2, and \#3, as well as the ground truth (GT). Labels are: visible - \checkmark, not visible - x. The bottom row shows screenshots from the video. The Fleiss kappa agreement score is -0.2.}
    \label{fig:example_low_agreement}
\end{figure}

After spam removal, we compute the agreement score between the turkers using Fleiss kappa \cite{fleiss1973equivalence}. Over the entire  data set, the Fleiss agreement score is 0.35, indicating fair agreement. On the ground truth data, the Fleiss kappa score is 0.46, indicating moderate agreement. This fair to moderate agreement indicates that the task is difficult, and there are cases where the visibility of the  actions is hard to label. To illustrate, Figure \ref{fig:example_low_agreement} shows examples where the annotators had low agreement. %

Table \ref{tab:statistics} shows statistics for our final dataset of videos labeled with actions, and Figure 2 shows a sample video and transcript, with annotations.

\begin{table}
    \centering
    \scalebox{0.9}{
    \begin{tabular}{l r}
    \toprule
    Videos & 177  \\
    Video hours & 21 \\
    Transcript words & 302,316 \\
    Miniclips & 1,268 \\
    Actions & 14,769 \\
    Visible actions & 4,340 \\
    Non-visible actions & 10,429 \\
    \bottomrule
    \end{tabular}
    }
    \caption{Data statistics.}
    \label{tab:statistics}
\end{table}

For our experiments, we use the first eight YouTube channels from our dataset as train data, the ninth channel as validation data and the last channel as test data. Statistics for this split are shown in Table \ref{tab:train-test-eval-split}.

\begin{table}
    \centering
    \scalebox{0.9}{
    \begin{tabular}{l c c c}
    \hline
     & Train & Test & Validation\\
        \midrule
        \# Actions & 11,403 & 1,999 & 1,367\\
        \# Miniclips & 997 & 158 & 113 \\
        \# Actions/ Miniclip & 11.4 & 12.6 & 12.0 \\
    \hline 
    \end{tabular}
    }
    \caption{Statistics for the experimental data split.}
    \label{tab:train-test-eval-split}
\end{table}

\subsection{Discussion}
The goal of our dataset is to capture naturally-occurring, routine actions. Because the same action can be identified in different ways (e.g., ``pop into the freezer'', ``stick into the freezer"), our dataset has a complex and diverse set of action labels. These labels demonstrate the language used by humans in everyday scenarios; because of that, we choose not to group our labels into a pre-defined set of actions. Table \ref{tab:comparison_statistics} shows the number of unique verbs, which can be considered a lower bound for the number of unique actions in our dataset. On average, a single verb is used in seven action labels, demonstrating the richness of our dataset.

The action labels extracted from the transcript are highly dependent on the performance of the constituency parser. This can introduce noise or ill-defined action labels. Some acions contain extra words (e.g., ``brush my teeth of course''), or lack words (e.g., ``let me just''). Some of this noise is handled during the annotation process; for example, most actions that lack words are labeled as ``not visible'' or ``not an action'' because they are hard to interpret.

\section{Identifying Visible Actions in Videos}
Our goal is to determine if actions mentioned in the transcript of a video are visually represented in the video. We develop a multimodal model that leverages both visual and textual information, and we compare its performance with several single-modality baselines.

\subsection{Data Processing and Representations}\label{sec:datarepresentations}

Starting with our annotated dataset, which includes miniclips paired with transcripts and candidate actions drawn from the transcript, we extract several layers of information, which we then use to develop our multimodal model, as well as several baselines.

\mysubsection{Action Embeddings.}
To encode each action, we use both GloVe \cite{pennington2014glove} and ELMo \cite{peters2018deep} embeddings. 
When using GloVe embeddings, we represent the action as the average of all its individual word embeddings. We use embeddings with dimension 50. When using ELMo, we represent the action as a list of words which we feed into the default ELMo embedding layer.\footnote{Implemented as the ELMo module in Tensorflow} This performs a fixed mean pooling of all the contextualized word representations in each action. 

\mysubsection{Part-of-speech (POS).} 
We use POS information for each action. Similar to word embeddings \cite{pennington2014glove}, we train POS embeddings. We run the Stanford POS Tagger \cite{toutanova2003feature} on the  transcripts and assign a POS to each word in an action. To obtain the POS embeddings, we train GloVe on the Google N-gram corpus\footnote{http://storage.googleapis.com/books/ngrams/books/ datasetsv2.html} using POS information from the five-grams. Finally, for each action, we average together the POS embeddings for all the words in the action to form a POS embedding vector. 

\mysubsection{Context Embeddings.} 
Context can be helpful to determine if an action is visible or not. We use two types of context information, action-level and sentence-level. Action-level context  takes into account the previous action and the next action; we denote it as Context$_A$. These are each calculated by taking the average of the action's GloVe embeddings.
Sentence-level context considers up to five words directly before the action and up to five words after the action (we do not consider words that are not in the same sentence as the action); we denote it as Context$_S$. Again, we average the GLoVe embeddings of the preceding and following words to get two context vectors.

\mysubsection{Concreteness.}
Our hypothesis is that the concreteness of the words in an action is related to its visibility in a video. 
We use a dataset of words with associated concreteness scores from \cite{brysbaert2014concreteness}. Each word is labeled by a human annotator with a value between 1 (very abstract) and 5 (very concrete). %
The percentage of actions from our dataset that have at least one word in the concreteness dataset is 99.8\%.
For each action, we use the concreteness scores of the verbs and nouns in the action. We consider the concreteness score of an action to be the highest concreteness score of its corresponding verbs and nouns.  \cref{tab:concr1}  shows several sample actions along with their concreteness scores and their visiblity. 

\begin{table}
\scalebox{0.95}{
    \begin{tabular}{l c c}
    \hline 
    Action & Con. & Visible? \\
        \midrule
        cook things in \textbf{water}&5.00&\checkmark\\
        head right into my \textbf{kitchen}&4.97&\checkmark\\
        throw it into the \textbf{washer}&4.70&\checkmark\\
        \midrule
        \textbf{told} you what& 2.31 & \xmark\\
        \textbf{share} my thoughts &2.96 & \xmark\\
        \textbf{prefer} them &1.62 & \xmark\\
    \hline 
    \end{tabular}
    }
    \caption{Visible actions with high concreteness scores (Con.), and non-visible actions with low concreteness scores. The noun or verb with the highest concreteness score is in bold.}
    \label{tab:concr1}
\end{table}

\begin{table}
\scalebox{0.95}{
    \begin{tabular}{l c}
    \hline 
    Action & Visible in the miniclip? \\
        \midrule
        put my son & \xmark\\
        sleep after we & \xmark\\
        done dinner & \xmark\\
        get comfortable & \checkmark\\
        pick out some pajamas& \checkmark\\
        start with my skincare & \xmark\\
        cleanse if I or even & \xmark\\
    \hline 
    \end{tabular}
    }
\end{table}

\mysubsection{Video Representations.} We use {\sc Yolo9000} \cite{redmon2017yolo9000} to identify objects present in each miniclip. We choose {\sc YOLO9000} for its high and diverse number of labels (9,000 unique labels). We sample the miniclips at a rate of 1 frame-per-second, and we use the {\sc Yolo9000} model pre-trained on COCO \cite{lin2014microsoft} and ImageNet \cite{deng2009imagenet}.

We represent a video both at the frame level and the sequence level. For frame-level video features, we use the Inception V3  model \cite{szegedy2016rethinking} pre-trained on ImageNet. We extract the output of the very last layer before the Flatten operation (the ``bottleneck layer"); we choose this layer because the following fully connected layers are too specialized for the original task they were trained for.
We extract Inception V3 features from miniclips sampled at 1 frame-per-second.

For sequence-level video features, we use the C3D model \cite{tran2015learning} pre-trained on the Sports-1M dataset \cite{karpathy2014large}. Similarly, we take the feature map of the sixth fully connected layer. Because C3D captures motion information, it is important that it is applied on consecutive frames. We take each frame used to extract the Inception features and extract C3D features from the 16 consecutive frames around it. 

We use this approach because combining Inception V3 and C3D features has been shown to work well in other video-based models \cite{jang2017tgif,carreira2017quo, kay2017kinetics}.

\subsection{Baselines}

Using the different data representations described in Section \ref{sec:datarepresentations}, we implement several baselines. 

\mysubsection{Concreteness.}
We label as visible all the actions that have a concreteness score above a certain threshold, and label as non-visible the remaining ones. We fine tune the threshold on our validation set; for fine tuning, we consider threshold values between 3 and 5. Table \ref{tab:all_results} shows the results obtained for this baseline.

\mysubsection{Feature-based Classifier.}
For our second set of baselines, we run a classifier on subsets of all of our features. We use an SVM \cite{cortes1995support}, and perform five-fold cross-validation across the train and validation sets, fine tuning the hyper-parameters (kernel type, C, gamma) using a grid search. 
We run experiments with various combinations of features: action GloVe embeddings; POS embeddings; embeddings of sentence-level context (Context$_S$) and action-level context (Context$_A$); concreteness score. The combinations that perform best during cross-validation on the combined train and validation sets are shown in Table \ref{tab:all_results}.

\begin{figure}[!t]
    \centering
    \includegraphics[width=\linewidth]{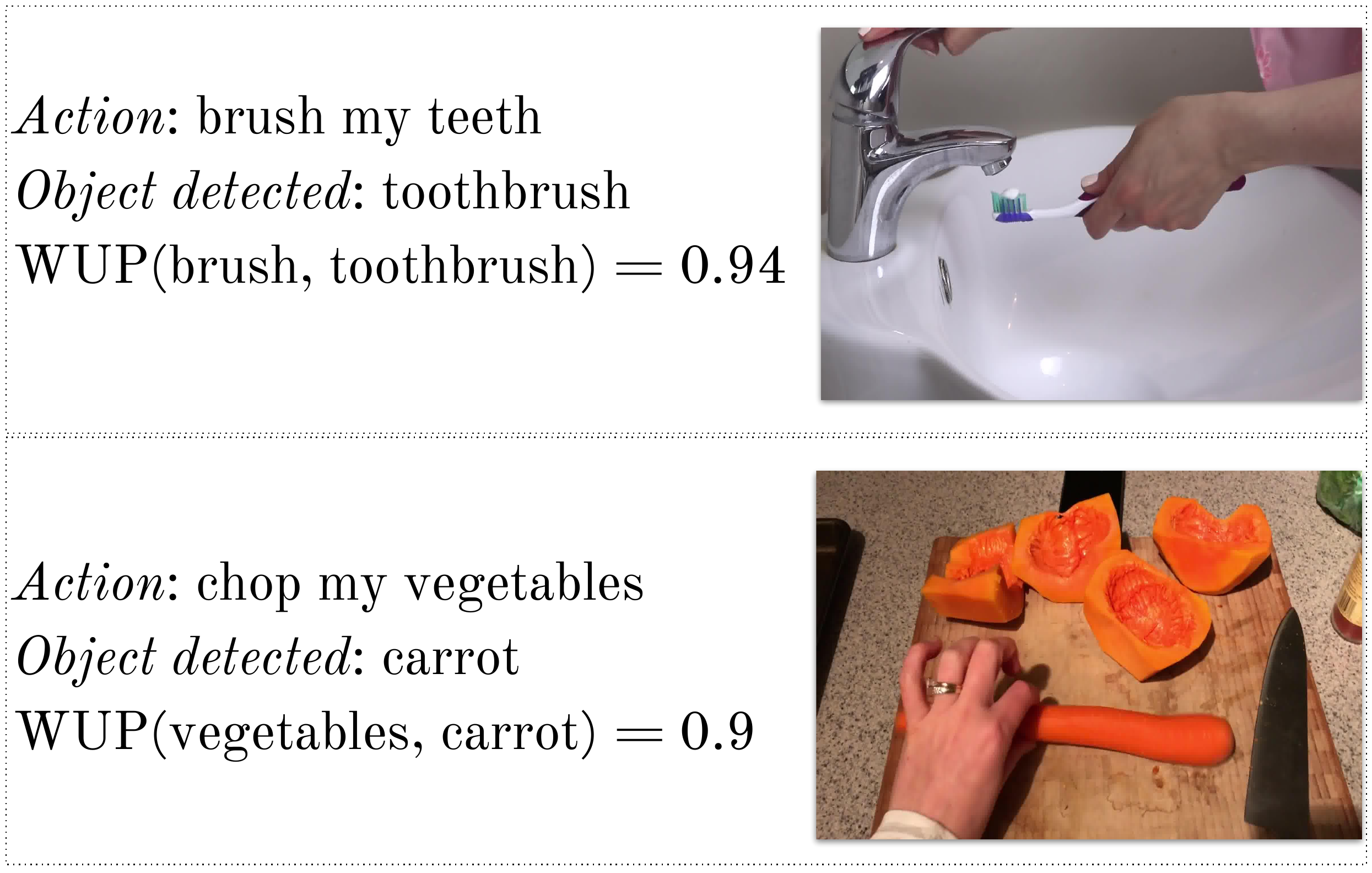}
    \caption{Example of frames, corresponding actions, object detected with {\sc Yolo}, and the object - word pair with the highest WUP similarity score in each frame. }
    \label{fig:yolo_example}
\end{figure}

\begin{figure*}[!t]
    \centering
    \includegraphics[width=\textwidth]{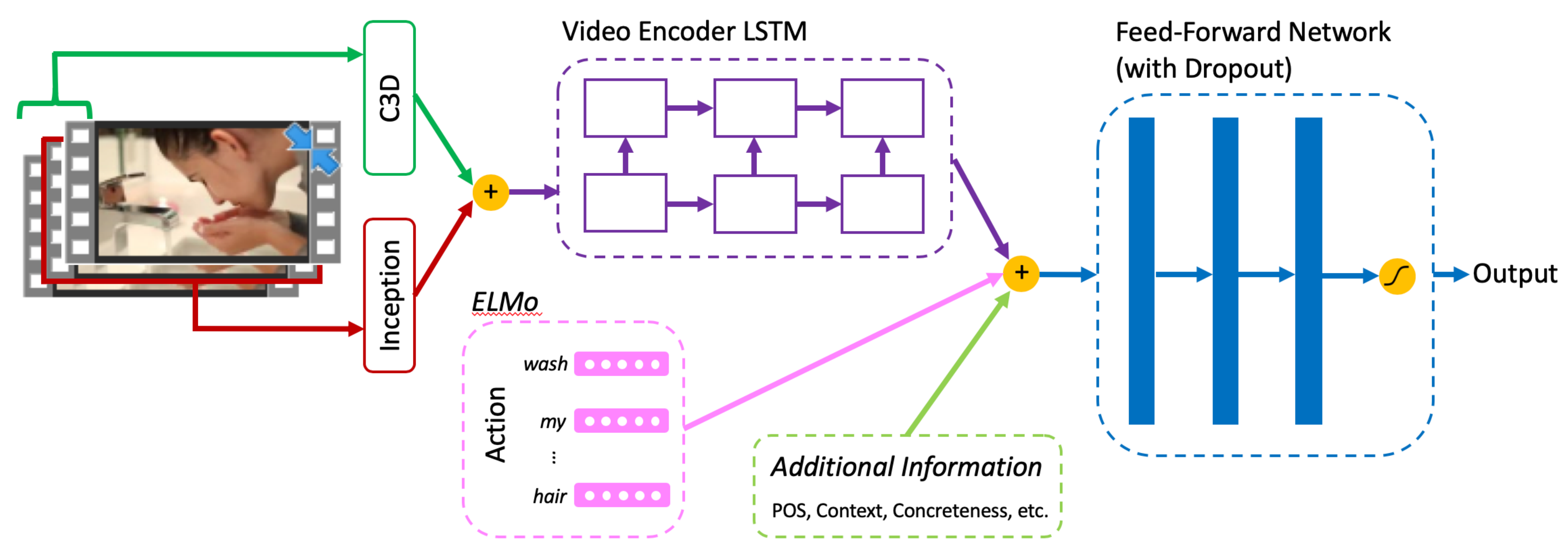}
    \caption{Overview of the multimodal neural architecture. + represents concatenation.}
    \label{fig:modelArchitecture}
\end{figure*}

\mysubsection{LSTM and ELMo.}
We also consider an LSTM model \cite{hochreiter1997long} that takes as input the tokenized action sequences padded to the length of the longest action. These are passed through a trainable embedding layer, initialized with GloVe embeddings, before the LSTM. The LSTM output is then passed through a feed forward network of fully connected layers, each followed by a dropout layer  \cite{srivastava2014dropout} at a rate of 50\%. We use a sigmoid activation function after the last hidden layer to get an output probability distribution. We fine tune the model on the validation set for the number of training epochs, batch size, size of LSTM, and number of fully-connected layers.

We build a similar model that embeds actions using ELMo (composed of 2 bi-LSTMs). We pass these embeddings through the same feed forward network and sigmoid activation function. The results for both the LSTM and ELMo models are shown in Table \ref{tab:all_results}. 

\mysubsection{{\sc Yolo} Object Detection.}
Our final baseline leverages video information from the {\sc YOLO9000} object detector. This baseline builds on the intuition that many visible actions involve visible objects. We thus label an action as visible if it contains at least one noun similar to objects detected in its corresponding miniclip. 
To measure similarity, we compute both the Wu-Palmer (WUP) path-length-based semantic similarity \cite{wu1994verbs} and the cosine similarity on the GloVe word embeddings. For every action in a miniclip, each noun is compared to all detected objects and assigned a similarity score.
As in our concreteness baseline, the action is assigned the highest score of its corresponding nouns. We use the validation data to fine tune the similarity threshold that decides if an action is visible or not. The results are reported in Table \ref{tab:all_results}. Examples of actions that contain one or more words similar to detected objects by {\sc Yolo} can be seen in Figure \ref{fig:yolo_example}.   %

\section{Multimodal Model}
Each of our baselines considers only a single modality, either text or video. While each of these modalities contributes important information, neither of them provides a full picture. The visual modality is inherently necessary, because it shows the visibility of an action. For example, the same spoken action can be labeled as either  \textit{visible} or \textit{non-visible}, depending on its visual context; we find 162 unique actions that are labeled as both visible and not visible, depending on the miniclip. This ambiguity has to be captured using video information. However, the textual modality provides important clues that are often missing in the video. The words of the person talking fill in details that many times cannot be inferred from the video. For our full model, we combine both textual and visual information to leverage both modalities.

We propose a multimodal neural architecture that combines encoders for the video and text modalities, as well as additional information (e.g., concreteness). Figure \ref{fig:modelArchitecture} shows our model architecture. The model takes as input a (miniclip $m$, action $a$) pair and outputs the probability that action $a$ is visible in miniclip $m$. We use C3D and Inception V3 video features extracted for each frame, as described in Section \ref{sec:datarepresentations}. %
These features are concatenated and run through an LSTM.

To represent the actions, we use ELMo embeddings (see Section \ref{sec:datarepresentations}). %
These features are  concatenated with the output from the video encoding LSTM, and run through a three-layer feed forward network with dropout. Finally, the result of the last layer is passed  through a sigmoid function, which produces a probability distribution indicating whether the action is visible in the miniclip. We use an RMSprop optimizer \cite{tieleman2012lecture} and fine tune the number of epochs, batch size and size of the LSTM and fully-connected layers.

\begin{table*}[!ht]
\scalebox{0.9}{
    \begin{tabular}{l|l|c c c c c c}
    \toprule
    Method & Input & Accuracy & Precision &  Recall &  F1\\
        \midrule
        \midrule
        \multicolumn{6}{c}{\sc Baselines} \\
        \midrule
        Majority & Action &0.692& 0.692 & 1.0 & 0.81 \\
        \midrule
        Threshold & Concreteness& 0.685 & 0.7 & 0.954 & 0.807 \\ %
        \midrule 
         &Action$_G$  &0.715 & 0.722 & 0.956 & 0.823 \\
\multirow{2}{5pt}{Feature-based Classifier} &Action$_G$, POS & 0.701 & 0.702 & \textbf{0.986} & 0.820 \\
         & Action$_G$, Context$_S$ & 0.725 & 0.736 & 0.938 & 0.825 \\
         &Action$_G$, Context$_A$ &0.712 & 0.722 & 0.949 & 0.820 \\
         &Action$_G$, Concreteness  &0.718 & 0.729 & 0.942 & 0.822 \\
         &Action$_G$, Context$_S$, Concreteness &\textbf{0.728} &
         \textbf{0.742} & 0.932 & \textbf{0.826} \\
        \midrule
        LSTM & Action$_G$ & 0.706 & 0.753 & 0.857 & 0.802 \\
        ELMo & Action$_G$ & \textbf{0.726} & \textbf{0.771} & \textbf{0.859} &\textbf{0.813} \\
        \midrule
        {\sc Yolo} & Miniclip  & 0.625 & 0.619 & 0.448& 0.520 \\
        \midrule
        \midrule
        \multicolumn{6}{c}{\sc Multimodal Neural Architecture (Figure \ref{fig:modelArchitecture})} \\
        \midrule 
          & Action$_E$, Inception & 0.722 & 0.765 &0.863 &0.811 \\
          & Action$_E$, Inception, C3D & 0.725 &0.769 &0.869 &0.814 \\
          & Action$_E$, POS, Inception, C3D  & 0.731 & 0.763 &0.885 &0.820 \\
\multirow{2}{30pt}{Multi-modal Model} & Action$_E$, Context$_S$, Inception, C3D & 0.725 & \textbf{0.770} &0.859 &0.812 \\
           & Action$_E$, Context$_A$, Inception, C3D   & 0.729 & 0.757 & 0.895 &0.820 \\
          & Action$_E$, Concreteness, Inception, C3D  &0.723 & 0.768& 0.860& 0.811 \\
          & Action$_E$, POS, Context$_S$, Concreteness, Inception, C3D & \textbf{0.737} & 0.758 &\textbf{0.911} &\textbf{0.827} \\
          \bottomrule
    \end{tabular}
    \caption{Results from baselines and our best multimodal method on validation and test data. Action$_G$ indicates action representation using GloVe embedding, and Action$_E$ indicates action representation using ELMo embedding. Context$_S$ indicates sentence-level context, and Context$_A$ indicates action-level context.}
    \label{tab:all_results}
}
\end{table*}

\section{Evaluation and Results}

Table \ref{tab:all_results} shows the results obtained using the multimodal model for different sets of input features. The model that uses all the input features available leads to the best results, improving significantly over the text-only and video-only methods.\footnote{Significance is measured using a paired t-test:   $p < 0.005$ when compared to the best text-only model; $p < 0.0005$ when compared to the best video-only model.} 

We find that using only {\sc Yolo} to find visible objects does not provide sufficient information to solve this task. This is due to both the low number of objects that {\sc Yolo} is able to detect,
and the fact that not all actions involve objects. For example, visible actions from our datasets such as ``get up", ``cut them in half", ``getting ready", and ``chopped up" cannot be correctly labeled using only object detection.  
Consequently, we need to use additional video information such as Inception and C3D information. 

In general, we find that the text information plays an important role. ELMo embeddings lead to better results than LSTM embeddings, with a relative error rate reduction of 6.8\%. This is not surprising given that ELMo uses two bidirectional LSTMs and has improved the state-of-the-art in many NLP tasks \cite{peters2018deep}. Consequently, we use ELMo in our multimodal model.

Moreover, the addition of extra information improves the results for both modalities. Specifically, the addition of context is found to bring improvements. The use of POS is also found to be generally helpful.

\section{Conclusion}

In this paper, we address the task of identifying human actions visible in online videos. We focus on the genre of lifestyle vlogs, and construct a new dataset consisting of 1,268 miniclips and 14,769 actions out of which 4,340 have been labeled as visible. We describe and evaluate several text-based and video-based baselines, and introduce a multimodal neural model that leverages visual and linguistic information as well as additional information available in the input data. We show that the multimodal model outperforms the use of one modality at a time.

A distinctive aspect of this work is that we label actions in videos based on the language that accompanies the video. This has the potential to create a large repository of visual depictions of actions, with minimal human intervention, covering a wide spectrum of actions that typically occur in everyday life. 

In future work, we plan to explore additional representations and architectures to improve the accuracy of our model, and to identify finer-grained alignments between visual actions and their verbal descriptions. 
The dataset and the code introduced in this paper are publicly available at \url{http://lit.eecs.umich.edu/downloads.html}.

\section*{Acknowledgments}
This material is based in part upon work supported by the Michigan Institute for Data Science, by the National Science Foundation (grant \#1815291), by the John Templeton Foundation (grant \#61156), and by DARPA (grant \#HR001117S0026-AIDA-FP-045). Any opinions, findings, and conclusions or recommendations expressed in this material are those of the author and do not necessarily reflect the views of the Michigan Institute for Data Science, the National Science Foundation, the John Templeton Foundation, or DARPA.

\bibliography{paper}
\bibliographystyle{acl/acl_natbib}

\end{document}